\pdfoutput=1

\documentclass[11pt]{article}
\usepackage[final]{acl}

\usepackage{times}
\usepackage{latexsym}
\usepackage[most]{tcolorbox}

\usepackage{pifont} 
\newcommand{\cmark}{\ding{51}} 
\newcommand{\xmark}{\ding{55}} 
\usepackage{booktabs}
\usepackage{multirow}
\usepackage{xcolor}
\usepackage{colortbl}
\usepackage{array} 
\usepackage{amsmath}
\usepackage{xcolor}
\usepackage{makecell}
\usepackage{subcaption}

\usepackage{tabularx}

\newcolumntype{C}{>{\centering\arraybackslash}p{0.07\textwidth}}
\usepackage[T1]{fontenc}

\usepackage[utf8]{inputenc}

\usepackage{microtype}

\usepackage{inconsolata}

\usepackage{graphicx}
\usepackage[ruled,linesnumbered]{algorithm2e} 
\usepackage{amsmath}
\usepackage{etoolbox} 
\title{
StrucSum: Graph-Structured Reasoning for Long Document Extractive Summarization with LLMs
}

\author{
  Haohan Yuan\textsuperscript{1} \quad
  Sukhwa Hong\textsuperscript{2} \quad
  Haopeng Zhang\textsuperscript{1} \thanks{corresponding author} \\
  \textsuperscript{1}ALOHA Lab, University of Hawaii at Manoa \\
  \textsuperscript{2}University of Hawaii at Hilo \\
  \texttt{\{haohany, sukhwa, haopengz\}@hawaii.edu} 
}

\begin{document}
\maketitle


\begin{abstract}
Large language models (LLMs) have shown strong performance in zero-shot summarization, but often struggle to model document structure and identify salient information in long texts. In this work, we introduce StrucSum, a training-free prompting framework that enhances LLM reasoning through sentence-level graph structures. StrucSum injects structural signals into prompts via three targeted strategies: Neighbor-Aware Prompting (NAP) for local context, Centrality-Aware Prompting (CAP) for importance estimation, and Centrality-Guided Masking (CGM) for efficient input reduction. Experiments on ArXiv, PubMed, and Multi-News demonstrate that StrucSum consistently improves both summary quality and factual consistency over unsupervised baselines and vanilla prompting. In particular, on ArXiv, it increases FactCC and SummaC by 19.2\%  and 8.0\% points, demonstrating stronger alignment between summaries and source content. The ablation study shows that the combination of multiple strategies does not yield clear performance gains; therefore, structure-aware prompting with graph-based information represents a promising and underexplored direction for the advancement of zero-shot extractive summarization with LLMs. Our source code is publicly available. \footnote{\url{https://github.com/HaohanYuan01/StrucSum}}

\end{abstract}

\section{Introduction}



Despite recent advances in LLM-based extractive summarization~\cite{zhang2023extractive}, summarizing long and structured documents remains challenging for large language models due to limited context windows and insufficient discourse awareness~\cite{zhang2024systematicsurveytextsummarization}. As documents increase in length and complexity, effective summarization requires not only identifying salient sentences but also reasoning over discourse structure and capturing long-range dependencies~\cite{zhong-litman-2025-discourse, 2022EmpiricalSurveyonLongDocument}, which continue to pose difficulties for existing approaches.
\begin{figure}[t]
  \centering
  \includegraphics[width=\linewidth]{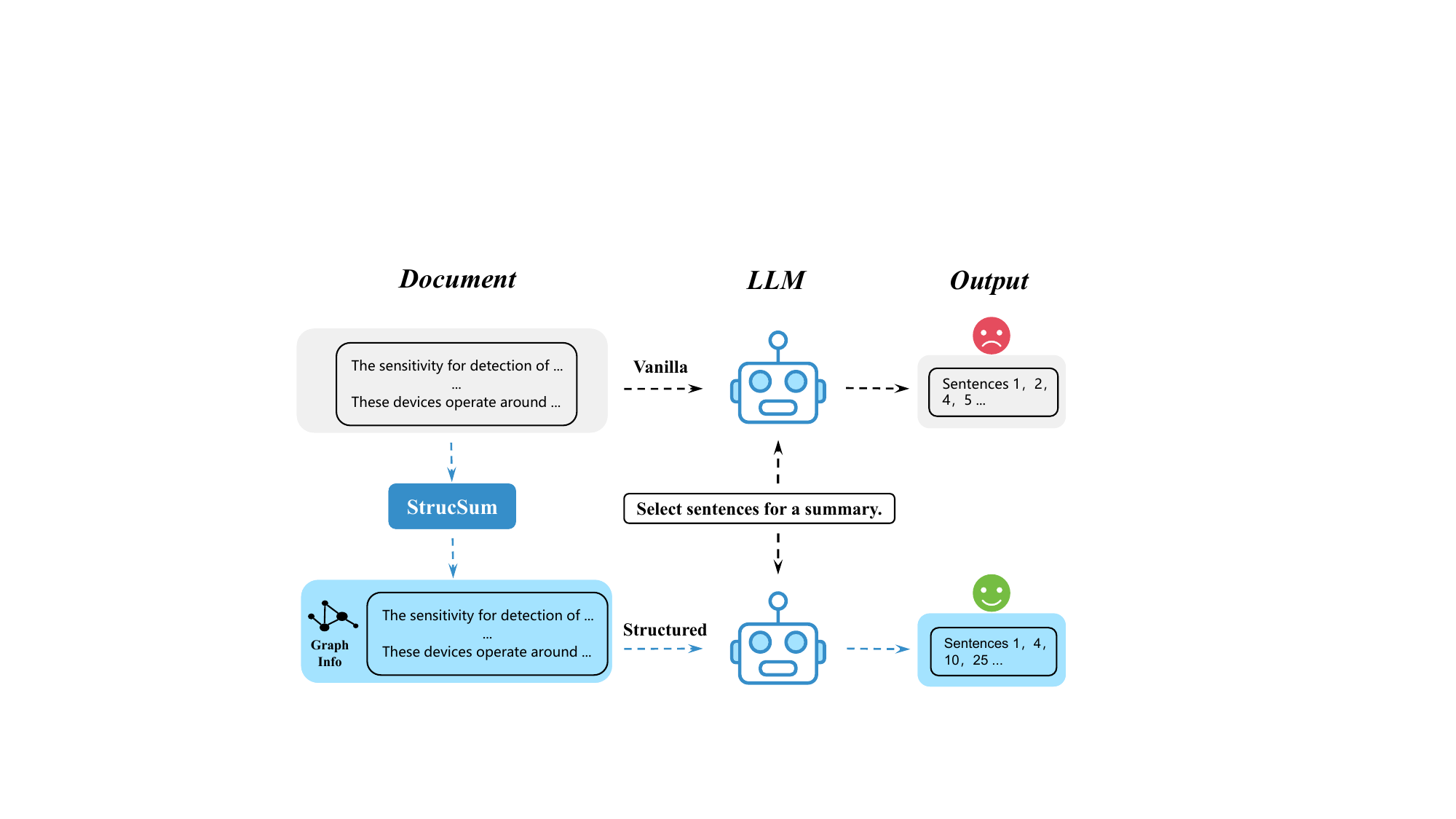}
  \caption{Comparing StrucSum's graph-structured LLM prompting with standard flat-input prompting for long-document extractive summarization.}

  \label{fig:graph_prompt}
\end{figure}

Extractive summarization selects key sentences to create concise, factually reliable summaries, crucial for domains requiring accuracy and traceability such as science, biomedicine, law, and policy.~\cite{zhang2023extractive, ruan2022histructimprovingextractivetext}.
To capture document-level structure, prior work has explored representing documents as sentence trees or graphs based on semantic, positional, or rhetorical relationships~\cite{mihalcea2004textrank, jin-etal-2020-multi, wang-etal-2020-heterogeneous}, offering interpretable mechanisms for sentence selection via centrality and coverage~\cite{Erkan:2004,cui-etal-2020-enhancing}. However, most existing approaches require supervised training with human written gold summaries~\cite{li-etal-2020-leveraging-graph, zhang-etal-2022-hegel, zhao2024hierarchical}, which limits their scalability and adaptability across domains.

Recently, LLMs have shown strong zero-shot performance in general summarization tasks~\cite{yuan2024domainsumhierarchicalbenchmarkfinegrained}. Their ability to generalize across domains without task-specific fine-tuning makes them attractive alternatives to fully supervised models~\cite{kojima2023largelanguagemodelszeroshot}. However, LLMs are trained on unstructured text and often struggle to recognize salient content and maintain factual consistency in long-form documents due to their lack of structural awareness~\cite{zhang2024systematicsurveytextsummarization}. Although prompting strategies like in-context learning~\cite{brown2020language} and chain-of-thought~\cite{wei2023chainofthoughtpromptingelicitsreasoning} have enhanced LLM reasoning in other domains, how to incorporate structural reasoning in prompting for long-document summarization remains unexplored.

In this paper, we propose StrucSum, a structure-aware prompting method for long-document extractive summarization in a fully zero-shot setting with LLMs. StrucSum constructs a Text-Attributed Graph (TAG) where sentences are nodes linked by semantic relations, and introduces three strategies to inject structural signals into model inputs: (1) \textbf{Neighbor-Aware Prompting} (NAP) adds local sentence context; (2) \textbf{Centrality-Aware Prompting} (CAP) appends graph-based importance scores; and (3) \textbf{Centrality-Guided Masking} (CGM) removes low-salience content to reduce length. These strategies enhance structural reasoning without any parameter updates. While developed for extractive summarization, StrucSum is model-agnostic and can be readily extended to abstractive or multi-document settings. Across datasets and LLMs, StrucSum improves both summary quality and factuality, revealing that symbolic graph priors can guide black-box LLMs without training.
Our main contributions include:

\begin{itemize}
\item We propose StrucSum, a structure-aware prompting framework for extractive summarization in a zero-shot, training-free LLM setting, featuring three structural strategies: NAP, CAP, and CGM. \vspace{-5pt}
\item Experiments on three long-form summarization benchmarks show that StrucSum consistently improves both summary quality and factual consistency over standard prompting. \vspace{-5pt}
\item We find that different structural strategies support different goals: CGM is most effective for maximizing overlap with reference summaries, NAP best maintains factual consistency, and CAP offers a balance between performance and input length. These findings provide guidance for selecting strategies based on task needs.
\end{itemize} \vspace{-5pt}

\section{Related Work}

\subsection{Graph-based Summarization}

\begin{figure*}[t!]
    \centering
  
    \includegraphics[width=0.95\textwidth]{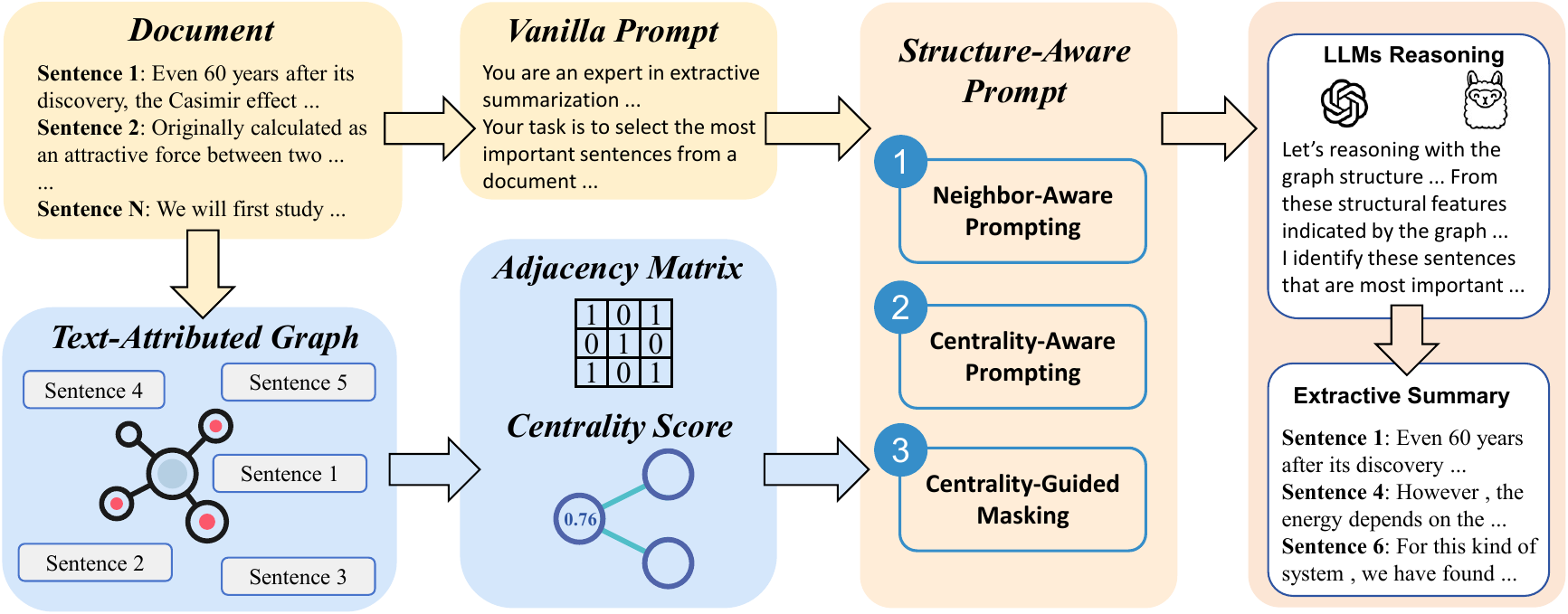}
    \caption{Overview of the StrucSum framework: A document is encoded into a Text-Attributed Graph, from which structural signals are extracted via three prompting strategies to guide LLMs in zero-shot extractive summarization.}
    \label{fig:hierarchies}
\end{figure*}

Graph-based models are a fundamental approach in extractive summarization, widely used for capturing cross-sentence relationships. Early work, such as TextRank~\cite{mihalcea2004textrank}, constructed sentence graphs based on lexical similarity to rank and select salient content. Subsequent research integrated Graph Neural Networks (GNNs) to learn more expressive, data-driven representations of these document structures~\cite{chhabra2024revisitingzeroshotabstractivesummarization}. This has led to the development of more complex graph designs for capturing finer-grained semantics, including the bipartite word-sentence graphs in MuchSUM~\cite{MuchSUM-mao-2022} and the joint modeling of words and sentences in heterogeneous GNNs~\cite{wang-etal-2020-heterogeneous}. These techniques have also been adapted for specific contexts: HGSUM~\cite{li2023compressed} uses a compressed heterogeneous graph for multi-document summarization, G-SEEK~\cite{moro2023graph} employs a lightweight selector for low-resource settings, and GSS constructs citation graphs for scientific articles~\cite{chen-etal-2022-scientific}.

\subsection{Summarization with LLMs}

Large Language Models have shown strong zero-shot performance on summarization benchmarks like CNN/DailyMail and XSum~\cite{zhang2024systematicsurveytextsummarization}. Despite these advances, a main challenge is factual consistency, as LLMs are prone to generating hallucinations unfaithful to the source content~\cite{pu2023summarizationalmostdead}. Furthermore, their effectiveness often decreases in specialized domains, such as medical or legal texts~\cite{yuan2024domainsumhierarchicalbenchmarkfinegrained}. To mitigate these issues, QA-prompting~\cite{sinha2025qapromptingimprovingsummarizationlarge} has been introduced to address positional biases by employing question-answering to extract key information before final summary generation. To address domain-specific difficulties, \citet{qi2025umbperanssumm2025enhancingperspectiveaware} improves summarization for medical texts by combining prompt optimization and supervised fine-tuning of LLMs. In addition, evaluation frameworks like G-Eval~\cite{liu2023gevalnlgevaluationusing,song2024finesurefinegrainedsummarizationevaluation} further enable finer-grained assessment of summary quality. Although some efforts explore training-free extractive summarization with LLMs~\cite{zhang2023extractive}, this area remains underdeveloped, particularly for long documents. Our work contributes to improving LLM reasoning in extractive summarization through zero-shot learning using structural signals, thereby eliminating the need for additional training or supervision.

\section{Method}

\subsection{Problem Definition}

In this work, we model zero-shot extractive summarization as a sentence selection task guided by structural reasoning. Each document is represented as a sequence of $N$ sentences, $\mathbf{S} = (s_1, s_2, \dots, s_N)$, and encoded as a Text-Attributed Graph (TAG): $\mathcal{G} = (\mathbf{V}, \mathcal{E})$, where each node $v_i \in \mathbf{V}$ corresponds to a sentence $s_i$, and each edge $(v_i, v_j) \in \mathcal{E}$ captures a semantic or positional relationship between sentence pairs. The objective is to select a subset of $k$ sentences from $\mathbf{S}$ to form the final summary $\hat{y} = \{s_{i_1}, s_{i_2}, \dots, s_{i_k}\}$.

Our goal is to construct a structured prompt $P$ that integrates both sentence content and graph structure, and use it to query a large language model (LLM) in a training-free setting. We formalize this in two steps:

\begin{equation}
P = GP(\mathbf{S}, \mathcal{G}),
\end{equation}
\begin{equation}
\hat{\mathcal{S}} = f(P). \quad \text{where } \hat{\mathcal{S}} \subseteq \mathbf{S}
\end{equation}

Here, $GP(\cdot)$ is a graph-aware prompting function that generates the structured input $P$ from the sentence set $\mathbf{S}$ and its associated graph $\mathcal{G} = (\mathbf{V}, \mathcal{E})$. The function $f(\cdot)$ denotes the LLM, which returns a subset of salient sentences $\hat{\mathcal{S}}$ as the extractive summary.

\subsection{Graph Construction}
\label{sec:graph_construction}

To capture inter-sentential relationships, we construct a sparse semantic graph for each document. Specifically, each sentence $s_i$ is encoded using a pre-trained Sentence-BERT model to obtain a dense vector. We compute cosine similarity between all sentence pairs $(s_i, s_j)$ and add an undirected edge $\mathbf{e}_{ij}$ to the edge set $\mathcal{E}$ if $\mathrm{sim}(s_i, s_j) > \theta$, where $\theta$ is a tunable similarity threshold. This construction yields a sparse graph structure, which preserves salient relational information while maintaining scalability for long documents.

Finally, this results in a TAG, $\mathcal{G} = (\mathbf{V}, \mathcal{E})$, which encodes both semantic proximity and discourse-level relationships among sentences. These TAGs provide structured signals, such as neighborhood context and sentence importance, integrated into the prompt design, allowing LLMs to reason over document structure during extraction.

\subsection{Vanilla Prompting}
\label{sec:vanilla_prompt}

We begin with a vanilla prompting strategy, which defines the basic mechanism for querying an LLM to perform extractive summarization. This prompt provides the LLM with an explicit task instruction and presents the document as a flat sequence of sentences, without incorporating any structural information. As such, it serves both as a fundamental way to elicit extractive behavior from the LLM and as a baseline for evaluating structure-aware extensions.

\paragraph{Task Description}
The LLM is instructed to identify and select the $k$ most important sentences from the document. The value of $k$ is a tunable hyperparameter that controls the target summary length.

\paragraph{Prompt Format}
The vanilla prompt defines the task and represents the document as an ordered list $\mathbf{S} = (s_1, s_2, \dots, s_N)$. This prompt template is reused and expanded upon in all structure-aware strategies.

\begin{figure}[!thb]
\centering
\fcolorbox{black}{gray!10}{
\parbox{0.92\linewidth}{
\small
\setlength{\baselineskip}{1.2\baselineskip}

\textbf{System Instruction:} \\
You are an expert in extractive summarization. \\
Your task is to \textbf{select the most important sentences} from a document. \\[0.2em]

\textbf{Guideline:} On average, select $k$ key sentences. \\[0.2em]

Sentence List: \\
Sentence 1: ``The UN Security Council met to discuss ...'' \\
Sentence 2: ``Several nations expressed concern over ...'' \\
Sentence 3: ``The meeting concluded with a call for ...'' \\
$\dots$ \\
Sentence N: ``International observers were invited to ...'' \\[0.2em]

\textbf{Expected Output Format:} \\
\texttt{\{ "selected\_sentences": [1, 3, 5] \}}
}
}
\vspace{-0.5em}
\label{fig:extractive-summary-prompt}
\end{figure}

\subsection{Structure-Aware Prompting}
\label{sec:prompting}

To better guide LLMs in reasoning over document structure, we propose three structure-aware prompting strategies that incorporate both local and global graph signals into the input. These strategies leverage sentence connectivity and importance derived from the TAG, enabling LLMs to make more informed extraction decisions in a fully training-free manner.

Each strategy reflects a different perspective on structural reasoning: 
(1) \textbf{Neighbor-Aware Prompting }integrates local sentence neighborhoods to enhance short-range discourse awareness; 
(2) \textbf{Centrality-Aware Prompting} emphasizes globally important sentences using graph centrality scores; and 
(3) \textbf{Centrality-Guided Masking} filters the input to include only structurally salient content, improving both reasoning focus and input efficiency.

The details of each prompting strategy are as follows:

\paragraph{Neighbor-Aware Prompting (NAP).}
NAP augments each sentence by explicitly listing its 1-hop neighbors in the TAG, allowing the LLM to consider short-range semantic dependencies and local discourse flow during selection.

The prompt augments each sentence as follows:

\begin{figure}[!thb]
\centering
\fcolorbox{black}{gray!10}{
\parbox{0.92\linewidth}{
\small
\setlength{\baselineskip}{1.2\baselineskip}

\textbf{Context:} Each sentence is followed by its 1-hop neighbors. \\[0.3em]

\textbf{Sentence 1:} ``\textless Text\_S1\textgreater'' \\
\textbf{Neighbors:} Sentence 2, 6, 7, 9, 13, 14 \\[0.2em]

\textbf{Sentence 2:} ``\textless Text\_S2\textgreater'' \\
\textbf{Neighbors:} Sentence 1, 4, 6, 9, 10, 15 \\
$\dots$ 
}
}
\label{fig:sentence-neighborhoods}
\end{figure}

By explicitly listing 1-hop neighbors, this format provides the LLM with local contextual cues that highlight semantically related sentences, helping it better assess the role of each sentence within its immediate discourse neighborhood.

\paragraph{Centrality-Aware Prompting (CAP).}
CAP emphasizes the global structural importance of each sentence by attaching a normalized degree centrality score, computed from the TAG. This score reflects how well-connected a sentence is within the document graph, highlighting key discourse hubs.

The centrality score for sentence $s_i$ is computed as:

\begin{equation}
\operatorname{centrality}(s_i) = \frac{1}{n - 1} \sum_{\substack{j=1 \\ j \neq i}}^{n} e_{ij} ,
\label{eq:centrality}
\end{equation}
where $e_{ij} = 1$ if an edge exists between $s_i$ and $s_j$, and $0$ otherwise.

The prompt format includes each sentence along with its centrality value:

\begin{figure}[thp]
\centering
\fcolorbox{black}{gray!10}{
\parbox{0.93\linewidth}{
\small
\setlength{\baselineskip}{1.15\baselineskip}

\textbf{Context:} Each sentence is presented with its centrality score. \\[0.1em]

\textbf{Sentence 1 (Centrality: 0.82)}: "\textless Text\_S1\textgreater" \\[0.2em]
\textbf{Sentence 2 (Centrality: 0.51)}: "\textless Text\_S2\textgreater" \\
$\dots$ 
}
}
\label{fig:cap-prompt}
\end{figure}

Including centrality values exposes global connectivity patterns, helping the LLM distinguish structurally prominent sentences from peripheral ones.

\paragraph{Centrality-Guided Masking (CGM).}
CGM introduces structural information by selectively including only a subset of high-centrality sentences in the prompt. Specifically, it ranks sentences by their normalized centrality scores and incrementally selects top-ranked ones until a cumulative centrality threshold $\rho$ (e.g., 85\%) is met. The remaining sentences are masked.

The following example shows the resulting prompt:

\begin{figure}[!thb]
\centering
\fcolorbox{black}{gray!10}{
\parbox{0.92\linewidth}{
\small
\setlength{\baselineskip}{1.1\baselineskip}

\textbf{Context:} Only top-ranked sentences (by centrality) are shown in full; others are masked. \\[0.2em]

\textbf{Sentence 1}: "\textless Text\_S1\textgreater" \\[0.2em]
\textbf{Sentence 4}: "\textless Text\_S4\textgreater" \\[0.2em]
\textbf{Sentence 6}: "\textless Text\_S6\textgreater" \\
$\dots$ 

}
}
\label{fig:cgm-prompt}
\end{figure}

The full procedure is detailed in Algorithm~\ref{alg:cgm}. This approach offers a more efficient encoding of structural relevance by reducing input length while preserving key informational content. Full prompt examples for vanilla prompting, NAP, CAP, and CGM are provided in the supplementary materials for clarity and reproducibility.

\begin{algorithm}[!ht]
\small
\KwIn{Sentences $\mathbf{S} = \{s_1, \dots, s_N\}$; Graph $\mathcal{G}$; Threshold $\rho \in (0,1)$}
\KwOut{Prompt $P$ with top-ranked sentences}
Compute centrality score $c_i$ for each $s_i$ using $\mathcal{G}$\;
Sort $\mathbf{S}$ by $c_i$ in descending order\;
Select minimal prefix $\{s_{(1)}, \dots, s_{(j)}\}$ such that $\sum_{k=1}^j c_{(k)} \geq \rho \cdot \sum_{i=1}^N c_i$\;
Build prompt $P$ with full text of selected sentences $\{s_{(1)}, \dots, s_{(j)}\}$\;
Mask the remaining sentences\;
\Return $P$\;
\caption{Centrality-Guided Masking (CGM)}
\label{alg:cgm}
\end{algorithm}

\section{Experiment}
\label{sec:experiment}

We conduct a systematic evaluation of our proposed structure-aware prompting strategies: Neighbor-Aware Prompting (NAP), Centrality-Aware Prompting (CAP), and Centrality-Guided Masking (CGM), for zero-shot extractive summarization with Large Language Models (LLMs).

\subsection{Experimental Setup}

\begin{table}
\small
\centering
\renewcommand{\arraystretch}{1.2}
\begin{tabular}{c | c | c | c}
\hline 
& \text{Arxiv} & \text{PubMed} & \text{MultiNews} \\
\hline
\# {test} & 200 & 200 & 200 \\
\text{document length} & 5,745 & 2,945 & 1,699 \\
\text{summary length} & 159 & 199 & 219 \\
\hline
\end{tabular}
\caption{Document and summary lengths (in words) for Arxiv, PubMed, and MultiNews datasets with 200 test samples.}
\label{data_stat}
\end{table}

\begin{table*}[t]
\setlength{\tabcolsep}{9pt}  
\renewcommand{\arraystretch}{1.1}
\centering
\begin{small}
\scalebox{1.05}{
\begin{tabular}{l|ccc|ccc|ccc}
\hline
& \multicolumn{3}{c|}{\textbf{\textsl{ArXiv}}} 
& \multicolumn{3}{c|}{\textbf{\textsl{PubMed}}} 
& \multicolumn{3}{c}{\textbf{\textsl{Multi-News}}} \\
\textbf{System} & \textbf{R-2} & \textbf{R-L} & \textbf{BS} & \textbf{R-2} & \textbf{R-L} & \textbf{BS} & \textbf{R-2} & \textbf{R-L} & \textbf{BS} \\
\hline

\rowcolor{gray!10}
\multicolumn{10}{c}{\textbf{\textsl{Unsupervised Systems}}} \\
\hline
\textsc{Lead} & 8.69 & 26.51 & 83.10 & 13.26 & 30.69 & 83.88 & 8.83 & 25.79 & 84.81 \\
LexRank~\citeyearpar{Erkan:2004} & 6.04 & 22.32 & 80.94 & 10.46 & 28.69 & 83.13 & 7.15 & 23.21 & 83.64 \\
TextRank~\citeyearpar{mihalcea2004textrank} & 8.69 & 23.37 & 81.63 & 12.20 & 29.27 & 83.29 & 9.32 & 25.57 & 84.35 \\
PacSum~\citeyearpar{zheng2019sentence} & 9.66 & 26.01 & 81.97 & 12.61 & 31.45 & 84.46 & 7.89 & 24.62 & 84.08 \\
\hline

\rowcolor{gray!10}
\multicolumn{10}{c}{\textbf{\textsl{Large Language Models}}} \\
\hline
GPT4o-mini
& 14.14 & 31.36 & 84.62 & 15.66 & 33.85 & 84.64 & 9.46 & 26.65 & 84.49 \\ 
+ NAP 
& 14.94 & 33.29 & 84.66 & 18.03 & 36.47& 84.67 & \cellcolor{red!18}10.13 & \cellcolor{red!18}27.63 & 84.77 \\ 
+ CAP
& \cellcolor{red!18}15.39 & 32.91 & \cellcolor{red!18}84.77 & 18.09 & 36.86 & 84.75 & 9.96 & 27.51 &  \cellcolor{red!18}84.94 \\ 
+ CGM
& 14.99 & \cellcolor{red!18}33.59 & 84.56 & \cellcolor{red!18}18.21 & \cellcolor{red!18}36.88 & \cellcolor{red!18}84.92 & 9.84 & 27.29 & 84.62 \\ 
\hline
GPT4o
& 13.74 & 31.87 & 84.57 & 17.53 & 36.00 & 85.19 & 9.49 & 26.97 & 84.67 \\ 
+ NAP
& \cellcolor{red!18}14.63 & 33.24 & 84.67 & 17.87 & 36.35 & 85.23 & 9.89 & 27.30 & \cellcolor{red!18} 84.70 \\ 
+ CAP
& 14.60 & \cellcolor{red!18}33.39 & \cellcolor{red!18}0.8478 & 17.91 & 36.73 & \cellcolor{red!18}85.25 & 9.84 & 27.33 & 84.69 \\ 
+ CGM
& 14.17 & 32.55 & 84.21 &\cellcolor{red!18}18.27 & \cellcolor{red!18}36.88 & \cellcolor{red!18}85.25 & \cellcolor{red!18}9.95 & \cellcolor{red!18}27.46 & 84.62 \\ 
\hline
LLaMA3-70B
& 13.67 & 31.32 & 82.68 & 16.14 & 34.21 & 85.21 & 10.38 & 27.87 & 84.28 \\ 
+ NAP
& \cellcolor{red!18}13.74 & \cellcolor{red!18}32.09 & \cellcolor{red!18}84.12 & \cellcolor{red!18}16.44 & \cellcolor{red!18}35.30 & 85.18 & 10.44 & 28.22 & \cellcolor{red!18}84.90 \\ 
+ CAP
& 13.62 & 31.69 & 83.09 & 16.18 & 34.72 & 0.85.14 & 10.34 & 27.76 & 84.74 \\
+ CGM
& 13.27 & 31.11 & 82.53 & 16.40 & 34.90 & \cellcolor{red!18}85.25 & \cellcolor{red!18}10.93 & \cellcolor{red!18}28.69 & 84.64 \\ 
\hline
\end{tabular}
}
\end{small}
\caption{Performance on ArXiv, PubMed, and Multi-News evaluated by ROUGE-2 (R-2), ROUGE-L (R-L), and BERTScore (BS). Best results within each LLM group are highlighted.}
\label{tab:results_rouge_bertscore}
\end{table*}
\paragraph{Datasets}
We conduct experiments on the following three summarization benchmarks:

\begin{itemize}
  \item \textbf{ArXiv}~\cite{cohan-etal-2018-discourse}: A long-document dataset of scientific papers from diverse domains, each paired with a structured abstract. \vspace{-20pt}

  \item \textbf{PubMed}~\cite{cohan-etal-2018-discourse}: A long-document dataset of biomedical research articles, each accompanied by a concise abstract summarizing complex findings. \vspace{-5pt}
  \item \textbf{Multi-News}~\cite{fabbri-etal-2019-multi}: A multi-document dataset containing clusters of news articles with human-written summaries that reflect diverse perspectives.
\end{itemize}

Due to the high inference cost of LLMs, we randomly sample 200 documents per dataset for development and testing. Average document and summary lengths (on the test set) are reported in Table~\ref{data_stat}.

\paragraph{Large Language Models}
We experiment with three LLMs: \textbf{GPT4o-mini} and \textbf{GPT4o} from OpenAI, and \textbf{LLaMA3.1-70B-Instruct} from Meta. GPT models are accessed via the official OpenAI API\footnote{\url{https://openai.com/index/openai-api/}}, while LLaMA3 is accessed through the NVIDIA API\footnote{\url{https://build.nvidia.com/models}}. For calling LLMs, we consistently set the temperature to $0$ and used the default ``top\_p'' value of $1.0$. The ``max\_tokens'' parameter was set to $100$ for all models to control the output length.

\vspace{5pt}

\paragraph{Evaluation}
We evaluate summary quality using a combination of standard, faithfulness-oriented, and holistic metrics. For content overlap, we report ROUGE-2 and ROUGE-L~\cite{lin-2004-rouge}, which capture n-gram overlap and longest common subsequence between system and reference summaries. For semantic similarity, we report BERTScore F1~\cite{zhang2020bertscoreevaluatingtextgeneration}, which measures contextual embedding similarity between summaries and references. To assess factual consistency, we use FactCC~\cite{kryściński2019evaluatingfactualconsistencyabstractive}, a classifier-based metric that detects factual errors in generated summaries, and SummaC~\cite{laban-etal-2022-summac}, which estimates sentence-level consistency using natural language inference models. Finally, we report G-Eval~\cite{liu2023gevalnlgevaluationusing}, a GPT-based evaluation framework that provides human-aligned scores across multiple dimensions, including coherence, consistency, fluency, and relevance. G-Eval offers a more comprehensive view of overall summary quality beyond traditional overlap-based metrics.

\paragraph{Unsupervised Baselines}  
We compare against widely-used unsupervised extractive summarization baselines. \textbf{Lead} selects the first few sentences as the summary, reflecting positional bias. \textbf{LexRank}~\cite{Erkan:2004} builds a similarity graph and ranks sentences via eigenvector centrality. \textbf{TextRank}~\cite{mihalcea2004textrank} uses a PageRank-like algorithm for ranking. \textbf{PacSum}~\cite{zheng2019sentence} incorporates position and polarity to assign edge weights, improving focus in sentence selection.

\paragraph{Implementation Details}
To construct the TAG, we first encode each sentence using the pre-trained \texttt{all-MiniLM-L6-v2} Sentence-BERT model. We then compute cosine similarity between all sentence pairs and add an undirected edge if their similarity exceeds a tunable threshold $\theta$. All hyperparameters are tuned on the development set. The values used for each dataset are provided in Appendix~\ref{sec:hyperparams}.

\subsection{Hyperparameter Settings}
\label{sec:hyperparams}

We report the values of three hyperparameters used in our framework. The settings are summarized in Table~\ref{tab:hyperparams}.

\begin{table}[h]
\centering
\renewcommand{\arraystretch}{1.2}
\setlength{\tabcolsep}{10pt} 
\small
\begin{tabular}{lccc}
\toprule
\textbf{Dataset} & $k$ & $\theta$ & $\rho$ \\
\midrule
PubMed      & 7 & 0.7 & 0.8 \\
ArXiv       & 7 & 0.6 & 0.8 \\
Multi-News  & 9 & 0.7 & 0.7 \\
\bottomrule
\end{tabular}
\caption{Hyperparameter values for number of selected sentences ($k$), similarity threshold for TAG construction ($\theta$), and centrality coverage ratio in CGM ($\rho$).}
\label{tab:hyperparams}
\end{table}

\subsection{Summary Quality Evaluation}
\label{main_result}

Table~\ref{tab:results_rouge_bertscore} reports the performance of StrucSum and baseline methods on three datasets: ArXiv, PubMed, and Multi-News. We compare unsupervised extractive baselines with LLM-based methods, including both vanilla prompting and structure-aware prompting. We observe that, in most cases, structure-aware prompting outperforms vanilla LLM prompting.

Among the baselines, Lead performs best on PubMed (R-2: 13.26) and remains competitive elsewhere. Graph-based methods like LexRank and PacSum show varying results but fall short of LLM performance, particularly on long documents.

For LLMs, all three structural strategies improve over vanilla prompting. On Multi-News, GPT4o-mini with NAP achieves an R-2 score of 10.13, representing the largest relative gain (+0.67) over its vanilla baseline. On ArXiv, CAP gives the highest R-2 score (15.39) and BERTScore (84.77) among GPT4o-mini variants. On PubMed, CGM performs best, with an R-2 of 18.21 and BERTScore of 84.92.

Overall, structure-aware prompting improves summary quality across different datasets. NAP provides the largest relative improvements, while CAP and CGM achieve the highest absolute scores on specific datasets and metrics.

\subsection{Summary Faithfulness Evaluation}
\label{sec:faithfulness_quality_eval}

Extractive summaries can still contain factual inconsistencies, such as coreference errors or misleading sentence combinations~\cite{zhang-etal-2023-extractivenotfaithful}. We evaluate summaries generated by GPT4o-mini on ArXiv and PubMed (Table~\ref{tab:factual_summary}) using three metrics: FactCC and SummaC for factual consistency, and G-Eval for summary evaluation with LLM-as-a-judge.

Across both datasets, all structure-aware strategies improve over vanilla prompting. NAP consistently achieves the highest scores on FactCC and SummaC, indicating its strong effectiveness in maintaining factual accuracy. It also slightly outperforms other methods on G-Eval. CAP and CGM show more metric-specific strengths: CAP performs competitively on G-Eval and FactCC, while CGM achieves strong gains on SummaC, particularly on PubMed.

We infer that NAP mainly reduces local reasoning errors rather than increasing reference overlap. By exposing each sentence’s 1-hop neighbors, NAP supplies a short-range discourse context that constrains the LLM’s sentence stitching. This directly lowers contradiction and attribution mistakes, which FactCC and SummaC are designed to detect, thereby yielding large gains on faithfulness-oriented metrics. 

These results suggest that structural prompting enhances the faithfulness of extractive summaries. Among the strategies, NAP is the most effective overall, while CAP and CGM offer complementary benefits depending on the focus of the evaluation. 

\begin{table}[t]
\small
\setlength{\tabcolsep}{3mm}
\centering
\resizebox{\linewidth}{!}{
\begin{tabular}{llccc}
\hline
\textbf{Dataset} & \textbf{System} & \textbf{FactCC} & \textbf{SummaC} & \textbf{G-Eval} \\
\hline

\multirow[c]{4}{*}{ArXiv} 
& GPT4o-mini  & 38.44  & 48.64  & 4.08  \\
& + NAP       & \textbf{57.67}  & \textbf{56.61}  & \textbf{4.23}  \\
& + CAP       & 45.26  & 49.07  & 4.14  \\
& + CGM       & 36.19  & 55.72  & 4.13  \\

\hline

\multirow[c]{4}{*}{PubMed} 
& GPT4o-mini  & 44.16  & 61.89  & 4.20  \\
& + NAP       & \textbf{55.44}  & \textbf{66.67}  & \textbf{4.32}  \\
& + CAP       & 49.43  & 65.56  & 4.25  \\
& + CGM       & 45.46  & 64.94 & 4.23  \\

\hline
\end{tabular}
}
\caption{
Evaluation results on summary factual consistency and holistic quality for GPT4o-mini with different prompting strategies. We report scores from two automated factual metrics (FactCC, SummaC) and one general-purpose LLM-based evaluator (G-Eval).
}
\label{tab:factual_summary}
\end{table}

\begin{table}[htb!]
\small
\setlength{\tabcolsep}{2.5mm}
\centering
\begin{tabular}{c|c|c|c|c}
\hline
\textbf{NAP} & \textbf{CAP} & \textbf{CGM} & \textbf{ROUGE} & \textbf{SummaC} \\
\hline
\xmark & \xmark & \xmark & 0.2781 & 0.5527 \\
\cmark & \xmark & \xmark & \underline{0.3005} & \textbf{0.6164} \\
\xmark & \cmark & \xmark & 0.2998 & 0.5732 \\
\xmark & \xmark & \cmark & \textbf{0.3034} & 0.6033 \\
\cmark & \cmark & \xmark & 0.2974 & 0.5893 \\
\xmark & \cmark & \cmark & 0.2979 & 0.5890 \\
\cmark & \xmark & \cmark & 0.3002 & \underline{0.6160} \\
\cmark & \cmark & \cmark & 0.3002 & 0.5908 \\
\hline
\end{tabular}
\caption{\label{tab:pubmed_joint_full}
Ablation and joint analysis of structure-aware strategies on the average of PubMed and ArXiv using GPT4o-mini. ROUGE (average of ROUGE-1/2/L) and SummaC are reported.
}

\end{table}
\subsection{Ablation Study of Structural Strategies}

We evaluate all valid combinations of NAP, CAP, and CGM on the average of PubMed and ArXiv using GPT-4o-mini. As shown in Table~\ref{tab:pubmed_joint_full}, individual strategies generally perform better than combined ones. CGM alone achieves the highest ROUGE score, while NAP alone gives the highest SummaC score. This suggests that CGM is more helpful for improving content coverage, whereas NAP is more effective for consistency. The full combination of NAP, CAP, and CGM performs in the middle for both metrics. In most cases, combining strategies does not lead to further improvement. These results indicate that each structural strategy focuses on different aspects of the summarization task, and using them together does not necessarily lead to better overall performance.

\subsection{Human Evaluation}
\begin{table}[t]
\centering
\setlength{\tabcolsep}{4.5pt}
\renewcommand{\arraystretch}{1.15}
\begin{small}
\begin{tabular}{lcccc}
\toprule
\textbf{Method} & \textbf{Fluent} & \textbf{Informative} & \textbf{Faithful} & \textbf{Overall} \\
\midrule
GPT4o-mini           & 2.47            & 3.81                 & 3.61              & 3.30             \\
+ NAP            & \textbf{2.92}   & \textbf{3.99}        & \textbf{3.95}     & \textbf{3.74}    \\
+ CAP            & 2.57            & 3.80                 & 3.70              & 3.38             \\
+ CGM            & 2.68            & 3.86                 & 3.79              & 3.51             \\
\bottomrule
\end{tabular}
\end{small}
\caption{Human evaluation (1–5 scale) on fluency, informativeness, faithfulness, and overall quality for GPT4o-mini with different prompting strategies.}
\label{tab:human_eval_single_col_updated}
\end{table}
We conducted a human evaluation on the PubMed dataset using a 1–5 scale for Fluency, Informativeness, Faithfulness, and Overall quality. As shown in Table \ref{tab:human_eval_single_col_updated}, all proposed strategies (NAP, CAP, CGM) outperform vanilla prompting, with NAP consistently achieving the highest scores across all criteria. The improvements in human-perceived Overall quality (e.g., NAP: 3.74 vs. Vanilla: 3.30) aligns with the gains observed in the holistic G-Eval metric (Table \ref{tab:factual_summary}). Furthermore, the improved human-rated Faithfulness (e.g., NAP: 3.95 vs. Vanilla: 3.61) is consistent with gains shown by automatic faithfulness metrics like FactCC and SummaC (Table \ref{tab:factual_summary}). The guidelines for our human evaluators are provided in Appendix~\ref{sec:human-eval-guidelines}.

\section{Analysis}



\subsection{Hyperparameter Sensitivity}
\label{sec:hyperparameter_sensitivity}
\begin{figure}[t]
  \centering
  \includegraphics[width=\linewidth]{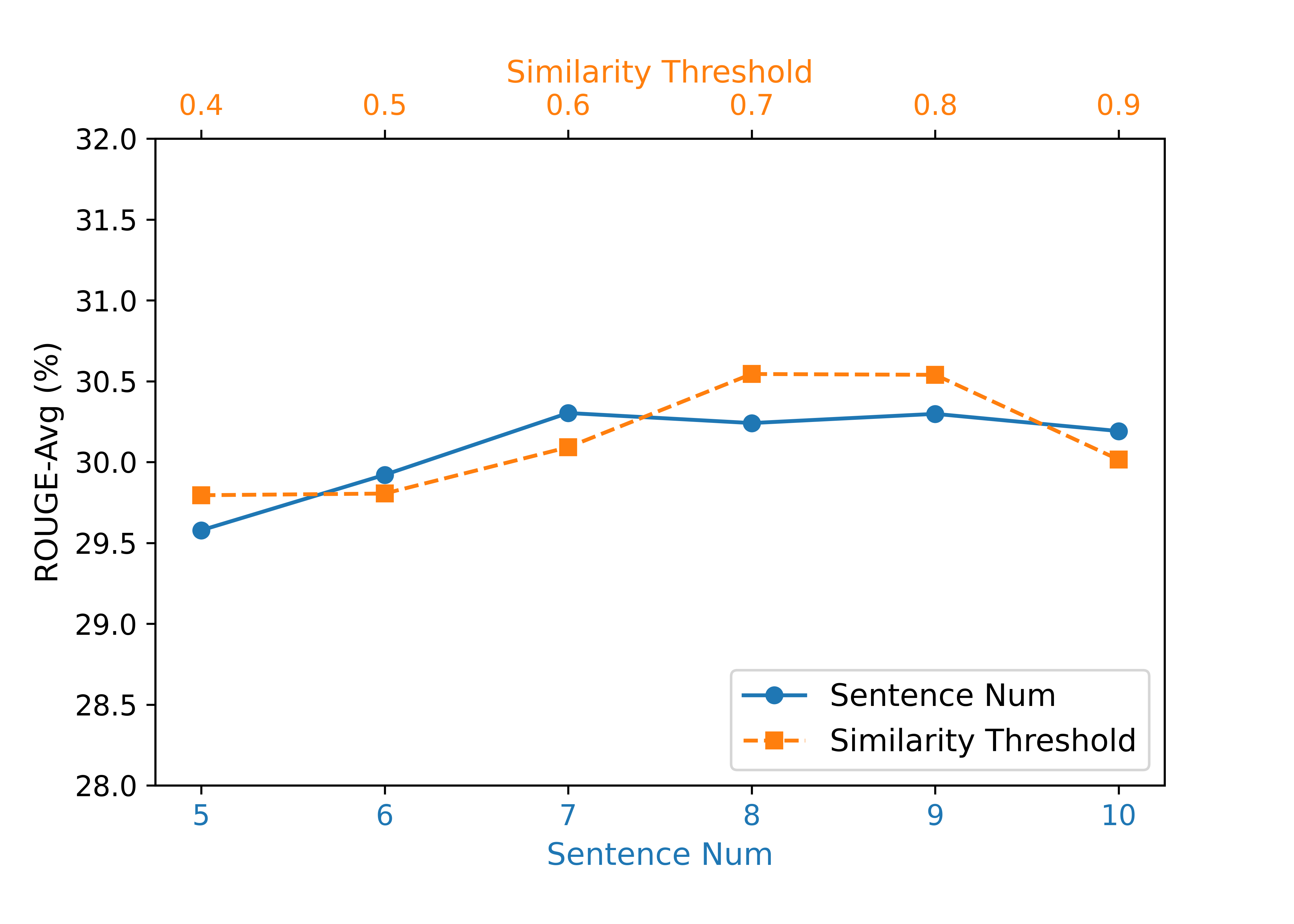} 
  \caption{Average ROUGE-F1 on PubMed with varying values of $k$ and $\theta$ in the NAP strategy.}
  \label{fig:hyper_sensitivity} 
\end{figure}

We evaluate the sensitivity of our proposed methods (we use the Neighbor-Aware Prompting method here as an example, while the other two strategies have shown similar results)  to two key hyperparameters: the target number of key sentences ($k$) and the graph construction similarity threshold ($\theta$). Figure~\ref{fig:hyper_sensitivity} illustrates these results, where different colored lines distinguish the ROUGE-Avg performance curves for each hyperparameter. As shown in Figure~\ref{fig:hyper_sensitivity}, the best ROUGE-Avg scores are achieved with approximately 7 key sentences ($k$) and a similarity threshold ($\theta$) between 0.7 and 0.8. Overall, the results suggest that our method achieves stable performance within a reasonable range of these hyperparameters.

\begin{figure}[t]
  \centering
  \includegraphics[width=\linewidth]{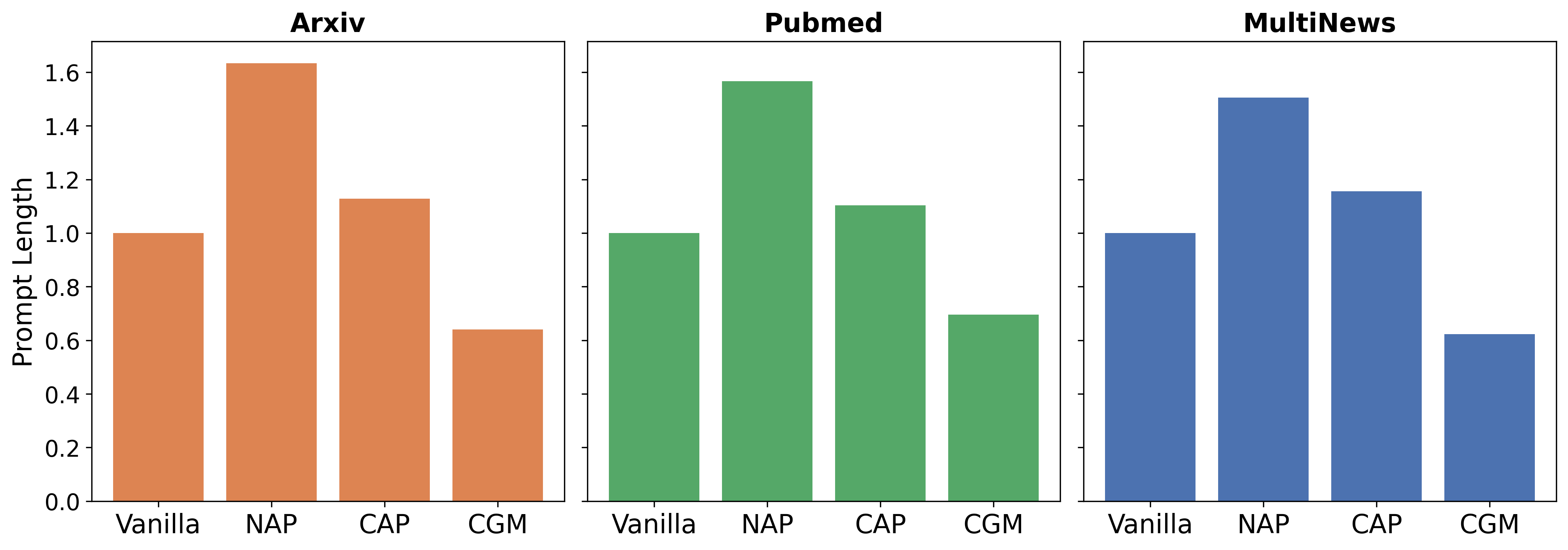}
  \caption{Normalized prompt lengths for vanilla, NAP, CAP, and CGM strategies across ArXiv, PubMed, and Multi-News.}
  \label{fig:prompt_length_comparison}
\end{figure}

\subsection{Token Usage Analysis}
\label{sec:token_usage_analysis}

Prompt token usage is a critical consideration for practical LLM applications~\cite{sahoo2025systematicsurveypromptengineering}. Figure~\ref{fig:prompt_length_comparison} shows the normalized prompt lengths for our structure-aware strategies: neighbor-aware prompting (NAP), centrality-aware prompting (CAP) and centrality-guided masking (CGM) relative to the vanilla prompting method. Encoding more structural context, NAP and CAP increase input length by approximately 50–60\% and 15–18\%, respectively. In contrast, CGM, reduces token usage by 40–50\% while still showing performance gains over the baseline methods (as detailed in Sections~\ref{main_result} and~\ref{sec:faithfulness_quality_eval}). Therefore,  CGM can be considered as an efficient option for scenarios constrained by input budget or latency.

\begin{figure}[t]
  \centering
  \includegraphics[width=\linewidth]{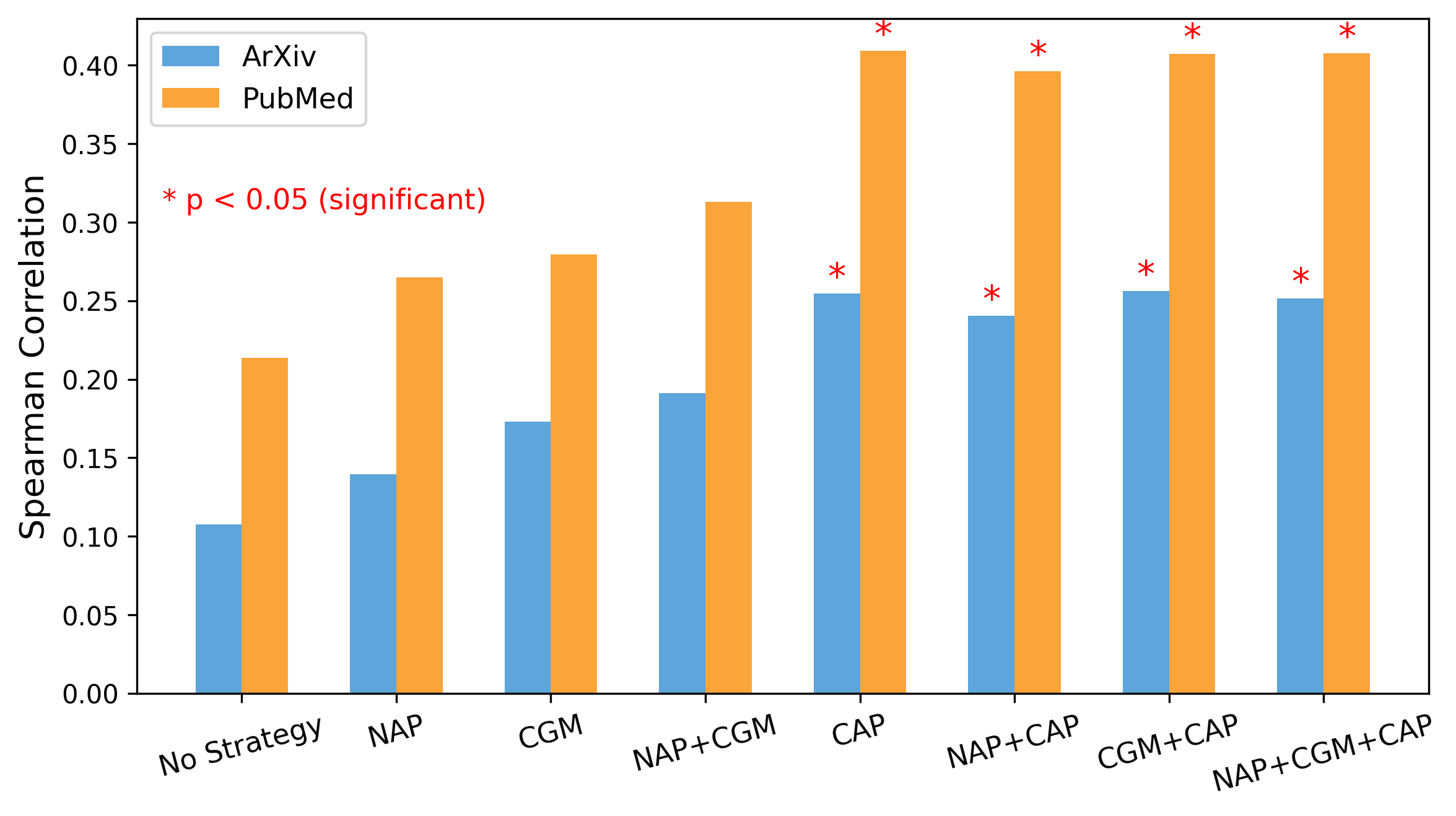}
  \caption{
Spearman correlation coefficients between sentence centrality scores and selection rank under different prompting strategies on GPT-4o, shown separately for ArXiv and PubMed. Asterisks indicate statistical significance ($p < 0.05$).
}

  \label{fig:correlation_analysis}
\end{figure}
\subsection{Correlation Analysis}

We assess whether LLMs follow structural cues by computing Spearman correlations between sentence centrality and selection ranks. As shown in Figure~\ref{fig:correlation_analysis}, the prompts that incorporate centrality (via CAP or combined strategies) produce significant positive correlations ($p < 0.05$), indicating the alignment between selection behavior and importance based on graphs. In contrast, NAP and CGM alone show nonsignificant trends ($p > 0.05$), suggesting a limited effect without explicit importance signals. Although CAP successfully guides LLMs to prioritize central sentences, the results of Table~\ref{tab:pubmed_joint_full} show that CAP-based configurations achieve smaller gains in ROUGE and SummaC compared to NAP and CGM. We infer that this may be because overemphasizing centrality limits the model’s flexibility to consider other informative sentences, which can affect both content coverage and consistency.

\begin{table}[t]
\setlength{\tabcolsep}{2.5pt}
\renewcommand{\arraystretch}{1.3}
\centering
\begin{small}\scalebox{1}{
\begin{tabular}{l|ccc|ccc}
\hline
& \multicolumn{3}{c|}{\textbf{\textsl{Arxiv}}} & \multicolumn{3}{c}{\textbf{\textsl{PubMed}}}\\
\textbf{System} & \textbf{R-2} & \textbf{R-L} & \textbf{BS}  & \textbf{R-2} & \textbf{R-L} & \textbf{BS}  \\
\hline
GPT4o-mini &  &  &  &  &  & \\
- TNL  & \textbf{11.87} & \textbf{30.05} & \textbf{83.91} & \textbf{13.33} & \textbf{32.03} & \textbf{84.44} \\
- NAM Matrix & 9.50 & 27.76 & 83.32 & 12.75 & 30.83 & 84.24 \\
- BAM  & 9.87 & 27.78 & 83.45 & 12.10 & 30.21 & 84.00 \\
\arrayrulecolor{black!30}\hline
+ NAP \textsuperscript{*} & 
14.59 & 33.15 & 84.66 & 16.42 & 35.42 & 85.28 \\
\arrayrulecolor{black}\hline
\end{tabular}}
\end{small}
\caption{
Evaluation results on ArXiv and PubMed under the graph-only setting using GPT4o-mini. In this setting, all sentence text is removed and only graph structural information is provided as input. * Indicates results taken from Table~\ref{tab:results_rouge_bertscore}
.
}
\label{tab:results_zero}
\end{table}

\subsection{Structure-Only Prompt Format Analysis}
\label{sec:graph_repr_analysis}

To evaluate whether LLMs can utilize structural information without access to sentence content, we design a graph-only setting where all textual input is removed and only graph-derived structure is provided to GPT4o-mini. We compare three prompt formats that encode the same underlying Text-Attributed Graph (TAG) differently: textual neighbor lists (TNL), which list each sentence’s neighbors as in the NAP strategy but omit sentence text; numeric adjacency matrices (NAM), dense matrices with cosine similarity scores between sentence pairs; and binary adjacency matrices (BAM), which indicate edge presence based on a similarity threshold $\theta$.

As shown in Table~\ref{tab:results_zero}, TNL consistently outperforms both NAM and BAM across all metrics. However, its performance remains lower than systems combining structure-aware prompts with full sentence content (for example, NAP achieves 14.59 ROUGE-2 compared to 11.87 for TNL on ArXiv). Overall, these results suggest that in LLM-based summarization, structural cues, although helpful, cannot replace textual information to achieve optimal results.

\begin{table}[t]
\centering
\setlength{\tabcolsep}{4.5pt}
\renewcommand{\arraystretch}{1.15}
\begin{small}
\begin{tabular}{lccccc}
\toprule
\textbf{Dataset} & $\boldsymbol{\theta}$ & \textbf{Avg \#Nodes} & \textbf{Avg \#Edges} & \textbf{Density} \\
\midrule
\multirow{6}{*}{ArXiv}
 & 0.4 & 200 & 7792.6 & 0.1712 \\
 & 0.5 & 200 & 3197.6 & 0.0743 \\
 & 0.6 & 200 & 1076.5 & 0.0260 \\
 & 0.7 & 200 & 304.2  & 0.0072 \\
 & 0.8 & 200 & 85.6   & 0.0018 \\
 & 0.9 & 200 & 36.6   & 0.0007 \\
\midrule
\multirow{6}{*}{PubMed}
 & 0.4 & 83 & 2700.5 & 0.3011 \\
 & 0.5 & 83 & 1410.3 & 0.1627 \\
 & 0.6 & 83 & 607.5  & 0.0744 \\
 & 0.7 & 83 & 210.3  & 0.0274 \\
 & 0.8 & 83 & 62.6   & 0.0078 \\
 & 0.9 & 83 & 26.9   & 0.0027 \\
\bottomrule
\end{tabular}
\end{small}
\caption{Graph sparsity statistics averaged over 200 documents. }
\label{tab:graph_sparsity}
\end{table}

\paragraph{Graph Sparsity Analysis.}
We analyze the sparsity of the constructed similarity graphs for the ArXiv and PubMed datasets under different similarity thresholds $\theta \in \{0.4, 0.5, 0.6, 0.7, 0.8, 0.9\}$. 
Following standard definitions in network analysis~\cite{10.1093/acprof:oso/9780199206650.001.0001}, edge density is defined as $|E| / (|V|(|V|-1))$. 
As shown in Table~\ref{tab:graph_sparsity}, the resulting graphs are consistently sparse, and their densities decrease rapidly as the threshold increases (e.g., ArXiv: 0.1712 $\rightarrow$ 0.0007; PubMed: 0.3011 $\rightarrow$ 0.0027). Our experimental results confirm that the similarity graphs are sparse.


\section{Conclusion}

We propose StrucSum, a prompting-based framework for long-document extractive summarization that augments LLMs with structural signals from sentence-level graphs. StrucSum applies three strategies: namely, NAP, CAP, and CGM, to guide LLMs in a zero-shot setting. Experiments on three benchmarks show that structure-aware prompting improves both summary quality and factual consistency. 
StrucSum demonstrates that integrating simple graph priors into prompts enhances LLM reasoning over long documents. Future work may extend this framework to multilingual and dynamic graph reasoning.

\section*{Acknowledgment}
This work was supported in part by National Science Foundation awards 2201428, 2004014, and 2138259, as well as by computing resources provided through the NSF ACCESS projects CIS250940 and CIS24061. We thank the anonymous reviewers for their valuable suggestions.

\section*{Limitations}

Instead of conducting experiments on the entire test set, we evaluate a representative subset of 200 test examples per dataset due to budget limits. Previous research efforts have similarly tested LLMs on small subsets~\cite{goyal2023newssummarizationevaluationera, zhang-etal-2024-benchmarking}. As large-scale evaluations remain resource-intensive under current LLM APIs, we leave full-dataset testing to future work.

Our evaluation primarily relies on automatic metrics to assess summary quality and factual consistency. While we include human evaluation for support, its scale is limited. A broader human study, especially involving domain experts, may provide deeper insights.

StrucSum does not involve any fine-tuning, as it is designed for the zero-shot setting. This training-free formulation promotes generality and ease of deployment, though it may limit task-specific performance gains. Exploring fine-tuned variants remains an open direction.

We also restrict our comparisons to vanilla prompting and several unsupervised baselines, without evaluating against supervised or fine-tuned models. This design isolates the effects of structure-aware prompting in a consistent zero-shot context, though broader comparisons could offer further perspective.

Finally, our experiments are conducted in English and on a selected set of LLMs (e.g., GPT-4o, LLaMA3-70B). Evaluating StrucSum across languages, domains, and additional model families is left for future exploration.

\bibliography{custom}

\appendix
\newpage

\section{Prompt Examples}
\label{sec:structure_prompts}
In this section, we provide prompt examples for the four main methods evaluated in this paper: Vanilla Prompting and the three structure-aware strategies: NAP, CAP, and CGM.

\begin{figure}[ht]
\centering
\fcolorbox{black}{gray!10}{
\parbox{0.95\linewidth}{
\small
\setlength{\baselineskip}{1.15\baselineskip}

\textbf{System Instruction:} \\
You are an expert in extractive summarization. \\
Your task is to \textbf{select the most important sentences} from a document. \\[0.4em]

\textbf{Guideline:} On average, select $k$ key sentences. \\[0.4em]

\textbf{Sentence List:} \\
Sentence 1: ``The sensitivity for detecting trace gas molecules in air or other carrier gases by measuring the characteristic vibration–rotation absorption lines in the mid-infrared is far higher than in the near-infrared or visible range.'' \\[0.4em]
Sentence 2: ``Thus, highly efficient optoelectronic devices operating in the mid-infrared are essential for sensitive gas analysis and atmospheric pollution monitoring.'' \\[0.4em]
Sentence 3: ``Narrow-gap IV–VI semiconductors are promising candidates as mid-infrared detectors, and polycrystalline IV–VI photodetectors operating at room temperature are already commercially available.'' \\[0.4em]
$\dots$ \\
Sentence 29: ``At the low-wavelength edge of the stopband (2.47 $\mu$m), an interference fringe appears due to the total thickness of the structure.'' \\[0.4em]

\textbf{Expected Output Format:} \\
\texttt{\{ "selected\_sentences": [1, 3, 5] \}}\par
}
}
\caption*{\textbf{Vanilla Prompting}}
\end{figure}

\vspace{1em}

\begin{figure}[ht]
\centering
\fcolorbox{black}{gray!10}{
\parbox{0.95\linewidth}{
\small
\setlength{\baselineskip}{1.15\baselineskip}

\textbf{System Instruction:} \\
You are an expert in extractive summarization. \\
Your task is to \textbf{select the most important sentences} from a document. Use information about each sentence's neighboring sentences to better reason about local context. \\[0.4em]

\textbf{Guideline:} On average, select $k$ key sentences. \\[0.4em]

\textbf{Context:} Each sentence is followed by its 1-hop neighbors. \\[0.4em]

Sentence 1: ``The sensitivity for detecting trace gas molecules in air or other carrier gases by measuring the characteristic vibration–rotation absorption lines in the mid-infrared is far higher than in the near-infrared or visible range.'' \\
Neighbors: Sentence 2, 4, 7, 10, 13 \\[0.3em]

Sentence 2: ``Thus, highly efficient optoelectronic devices operating in the mid-infrared are essential for sensitive gas analysis and atmospheric pollution monitoring.'' \\
Neighbors: Sentence 1, 3, 5, 6, 11 \\[0.3em]

Sentence 3: ``Narrow-gap IV–VI semiconductors are promising candidates as mid-infrared detectors, and polycrystalline IV–VI photodetectors operating at room temperature are already commercially available.'' \\
Neighbors: Sentence 2, 4, 6, 8, 12 \\[0.3em]

$\dots$ \\[0.3em]

Sentence 29: ``At the low-wavelength edge of the stopband (2.47 $\mu$m), an interference fringe appears due to the total thickness of the structure.'' \\
Neighbors: Sentence 24, 26, 27, 28 \\[0.4em]

\textbf{Expected Output Format:} \\
\texttt{\{ "selected\_sentences": [1, 3, 5] \}}\par
}
}
\caption*{\textbf{Neighbor-Aware Prompting (NAP)}}
\end{figure}

\begin{figure}[ht]
\centering
\fcolorbox{black}{gray!10}{
\parbox{0.95\linewidth}{
\small
\setlength{\baselineskip}{1.15\baselineskip}

\textbf{System Instruction:} \\
You are an expert in extractive summarization. \\
Your task is to \textbf{select the most important sentences} from a document. Use the centrality scores provided to help identify globally important sentences. \\[0.4em]

\textbf{Guideline:} On average, select $k$ key sentences. \\[0.4em]

\textbf{Context:} Each sentence is presented with its centrality score. \\[0.4em]

Sentence 1 (Centrality: 0.82): ``The sensitivity for detecting trace gas molecules in air or other carrier gases by measuring the characteristic vibration–rotation absorption lines in the mid-infrared is far higher than in the near-infrared or visible range.'' \\[0.3em]
Sentence 2 (Centrality: 0.51): ``Thus, highly efficient optoelectronic devices operating in the mid-infrared are essential for sensitive gas analysis and atmospheric pollution monitoring.'' \\[0.3em]
Sentence 3 (Centrality: 0.65): ``Narrow-gap IV–VI semiconductors are promising candidates as mid-infrared detectors, and polycrystalline IV–VI photodetectors operating at room temperature are already commercially available.'' \\
$\dots$ \\[0.3em]
Sentence 29 (Centrality: 0.47): ``At the low-wavelength edge of the stopband (2.47 $\mu$m), an interference fringe appears due to the total thickness of the structure.'' \\[0.4em]

\textbf{Expected Output Format:} \\
\texttt{\{ "selected\_sentences": [1, 3, 5] \}}\par
}
}
\caption*{\textbf{Centrality-Aware Prompting (CAP)}}
\end{figure}

\begin{figure}[ht]
\centering
\fcolorbox{black}{gray!10}{
\parbox{0.95\linewidth}{
\small
\setlength{\baselineskip}{1.15\baselineskip}

\textbf{System Instruction:} \\
You are an expert in extractive summarization. \\
Your task is to \textbf{select the most important sentences} from a document. \\
The document has been pre-filtered to include only structurally salient sentences, identified via graph centrality. \\[0.4em]

\textbf{Guideline:} On average, select $k$ key sentences. \\[0.4em]

\textbf{Context:} Only top-ranked sentences (by centrality) are shown in full; others are masked. \\[0.4em]

Sentence 1: ``The sensitivity for detecting trace gas molecules in air or other carrier gases by measuring the characteristic vibration–rotation absorption lines in the mid-infrared is far higher than in the near-infrared or visible range.'' \\[0.3em]
Sentence 4: ``This requirement can be met by using either a narrow-band emitter with a broadband detector, or a broadband source such as a glowbar with a detector of narrow spectral bandwidth.'' \\[0.3em]
Sentence 6: ``In this so-called resonant cavity enhanced photodetector (RCEPD), a thin absorbing layer is placed in a vertical optical cavity.'' \\
$\dots$ \\[0.3em]
Sentence 29: ``At the low-wavelength edge of the stopband (2.47 $\mu$m), an interference fringe appears due to the total thickness of the structure.'' \\[0.4em]

\textbf{Expected Output Format:} \\
\texttt{\{ "selected\_sentences": [1, 3, 5] \}}\par
}
}
\caption*{\textbf{Centrality-Guided Masking (CGM)}}
\end{figure}

\onecolumn

\clearpage
\section{Human Evaluation Guidelines}
\label{sec:human-eval-guidelines}
Our human evaluation used two PhD students, each independently rating summaries using the rubric on a 1–5 scale for Fluency, Informativeness, Faithfulness, and Overall Quality. We report the average of their scores across examples. 
Here we list the full human evaluation guidelines used for our manual annotation study.
including rating scales and definitions.

\begin{table*}[h!]
\small
\centering
\renewcommand{\arraystretch}{1.15}
\label{tab:human-eval-rubric}
\begin{tabularx}{\textwidth}{cXXXX}
\toprule
\textbf{Score} & \textbf{Fluency} & \textbf{Informativeness} & \textbf{Faithfulness} & \textbf{Overall Quality} \\
\midrule
\textbf{5} &
Grammatical, natural, and well-structured; reads smoothly with appropriate register; no typos or awkward phrasing. &
Covers all salient points and key details from the source; concise with minimal redundancy; no important omissions. &
Fully faithful to the source; no hallucinations, distortions, or unsupported inferences; entity, number, and event details are correct. &
Publication-quality summary: accurate, concise, coherent, and follows instructions; no substantive issues. \\
\midrule
\textbf{4} &
Mostly fluent with only minor issues (rare typos or slightly awkward phrasing) that do not impede comprehension. &
Covers most key points, with at most minor omissions or inclusion of unimportant details. &
Largely faithful; at most very minor imprecision that does not change meaning or introduce new facts. &
High-quality summary with minor issues in at most one dimension; clearly useful as-is. \\
\midrule
\textbf{3} &
Generally readable but with noticeable issues (grammar/style/word choice) that occasionally hinder the flow. &
Conveys several main points but misses multiple important details or includes some irrelevant/redundant content. &
Some unsupported statements, paraphrase drift, or partial contradictions; localized factual issues are present. &
Adequate but requires edits; there is clear room for improvement in two or more dimensions. \\
\midrule
\textbf{2} &
Frequent grammatical errors, unclear or choppy sentences; difficult to follow at times. &
Captures few key points; omits many important details and/or contains substantial irrelevant content. &
Multiple factual errors, contradictions, or hallucinated content; the meaning often diverges from the source. &
Low-quality summary with substantial issues; not reliable without significant revision. \\
\midrule
\textbf{1} &
Severely ungrammatical or incoherent; hard to understand. &
Fails to convey the main points; largely irrelevant or off-topic. &
Unfaithful to the source, with pervasive hallucinations or contradictions. &
Unusable summary that does not meet task requirements. \\
\bottomrule
\end{tabularx}
\caption{Human evaluation rubric (1--5 scale) for Fluency, Informativeness, Faithfulness, and Overall Quality, used to assess GPT-4o-mini summaries under different prompting strategies.}
\end{table*}

\end{document}